# Advancing Oyster Phenotype Segmentation with Multi-Network Ensemble and Multi-Scale mechanism

Wenli Yang[1], Yanyu Chen[1], Andrew Trotter[2], Byeong Kang[1]

1. School of ICT, College of Sciences and Engineering, University of Tasmania, Australia

2. Institute for Marine and Antarctic Studies, College of Sciences and Engineering, University

of Tasmania, Australia

**Abstract**

Phenotype segmentation is pivotal in analysing visual features of living organisms, enhancing our understanding of their characteristics. In the context of oysters, meat quality assessment is paramount, focusing on shell, meat, gonad, and muscle components. Traditional manual inspection methods are time-consuming and subjective, prompting the adoption of machine vision technology for efficient and objective evaluation. We explore machine vision's capacity for segmenting oyster components, leading to the development of a multi-network ensemble approach with a global-local hierarchical attention mechanism. This approach integrates predictions from diverse models and addresses challenges posed by varying scales, ensuring robust instance segmentation across components. Finally, we provide a comprehensive evaluation of the proposed method's performance using different real-world datasets, highlighting its efficacy and robustness in enhancing oyster phenotype segmentation.

1. Introduction

Phenotype segmentation, as a method to categorize and analyse visual features in living organisms, plays a crucial role in enhancing our understanding of their characteristics and variations. In the specific context of oysters, the quality assessment is primarily focused on the condition of their meat (all of the soft tissues). Swift and accurate evaluation of meat condition is essential to maximise the marketability oysters. This evaluation involves the identification and categorization of key oyster components of this phenotype, including the shell, meat, gonad, and muscle (adductor muscle) (Vu 2022), which collectively contribute to the overall quality and grading of oyster. These components shape both the visual and gustatory attributes highly valued by consumers.

Traditionally, the assessment of oyster meat condition has relied on manual inspection by seasoned experts. However, this approach is not only time-consuming but also subjective, leading to notable variations in grading and pricing (Wu et al. 2022). Recognizing the need for a more efficient and objective method, recent years have seen the integration of machine vision technology emerge as an innovative solution (Vo et al. 2021) for all aquaculture stages, like feeding (Daud et al. 2020), classification (Zhao et al. 2021), grading (Sung, Park, and Choi 2020) and counting (Cao and Xu 2018). These technologies offer notable advantages, including heightened accuracy, increased efficiency, and enhanced objectivity in the evaluation process.

At the core of this advanced assessment is the segmentation of oyster components—a crucial step involving the meticulous delineation and categorization of various elements within an image. The primary goal is to extract all pixels within the image and allocate them to specific component classes, such as the shell, meat, gonad, and muscle. The technology's capacity to objectively segment and categorize these components of oyster meat marks a significant stride forward in grading processes, ensuring a standardized and reliable evaluation of visual meat quality. This not only expedites the assessment process but also minimizes the potential for subjective discrepancies, reinforcing the commitment to delivering consistently high-quality oyster products.

In real-world scenarios, images capturing several oysters on a tray may vary in conditions, including different resolutions and varying numbers of oysters. A trade-off exists in this task, where certain types of detection and segmentation for specific components are better suited for lower inference resolutions, while others are more effective at higher inference resolutions. Examples and investigations of these cases are presented in Table 1. In this investigation, we noted a progressive rise in complexity across the three chosen backbones for the Mask-RCNN model (He et al. 2017), progressing from ResNet-50, ResNet-101 (He et al. 2016) and ResNeXt (Xie et al. 2017). This progression was undertaken to assess the model's performance on images of different scales, aiming specifically at the instance segmentation of four distinct components: shell, meat, gonad, and muscle.

Table 1. Investigations among different models and scales (IoU=0.6).

|  |  | Shell | Meat | Gonad | Muscle |
|---|---|---|---|---|---|
| Scale 2.0 | R-50 | 87.00% | 77.02% | 51.16% | 52.56% |
|  | R-101 | 88.07% | 70.70% | 56.23% | 48.81% |
|  | ResNeXt | 85.49% | 73.74% | 66.13% | 48.46% |
| Scale 1.0 | R-50 | 86.99% | 74.81% | 51.35% | 48.86% |
|  | R-101 | 87.34% | 71.08% | 57.19% | 49.38% |
|  | ResNeXt | 85.51% | 77.51% | 66.23% | 54.15% |
| Scale 0.5 | R-50 | 87.73% | 73.82% | 53.91% | 47.95% |
|  | R-101 | 87.41% | 69.03% | 53.26% | 52.91% |
|  | ResNeXt | 86.09% | 76.67% | 65.98% | 50.98% |

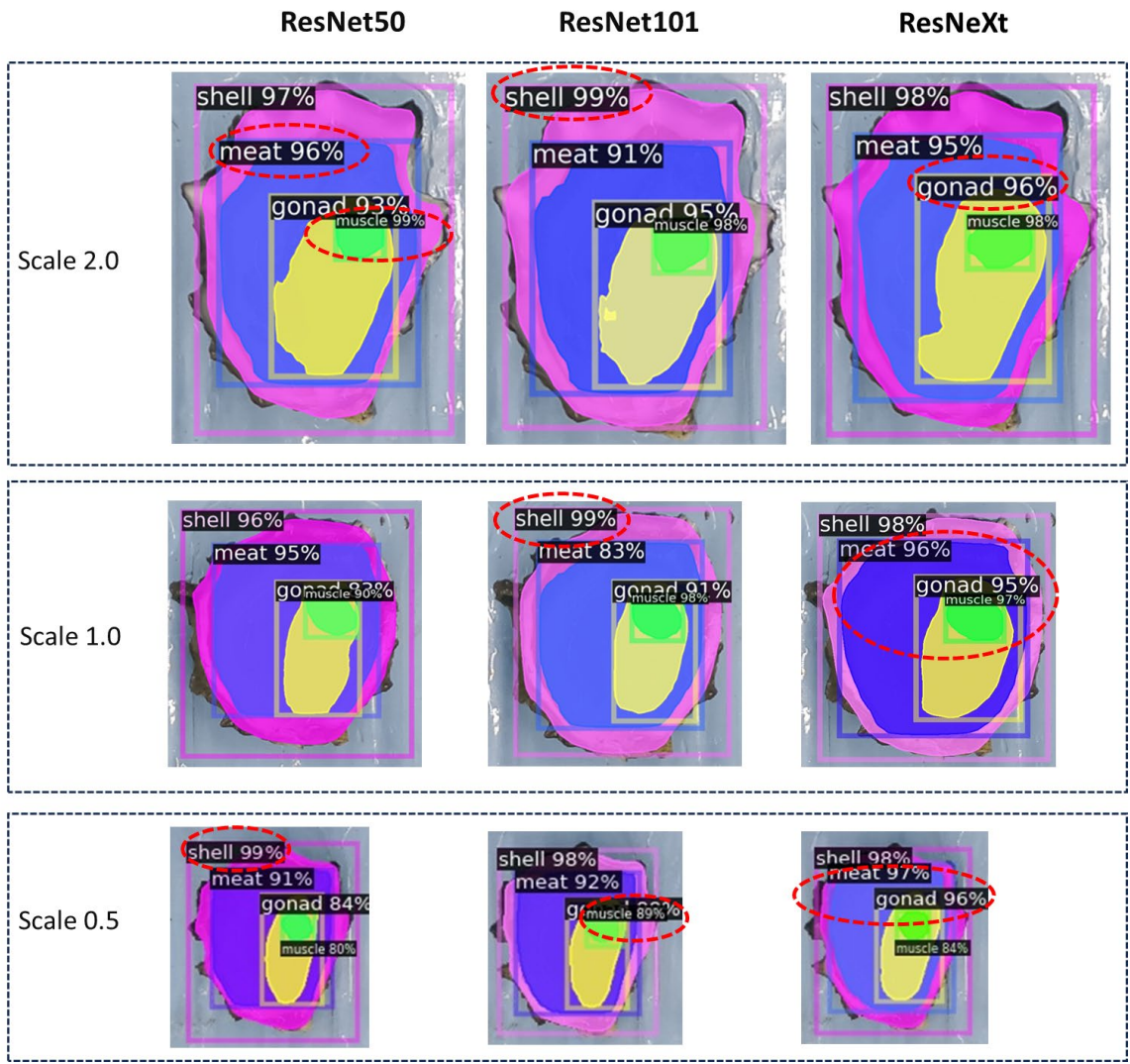

Fig.1. Sample segmentation results among different models and scaled images (visually representing the segmentation of four distinct components – shell, meat, gonad, and muscle and each component is assigned custom colours for clarity).

Illustrated in Table 1 an Fig. 1, the results depict various instance segmentation modes concerning the inference scale. In the upscaled images, ResNet50 demonstrates superior segmentation for the meat and muscle components, whereas RetNet101 and ResNeXt exhibit better segmentation for the shell and gonad, respectively. Conversely, in the downscaled images (third-row low-resolution images), ResNet50 and ResNet101 exhibit enhanced performance in shell and muscle segmentation, respectively, while ResNeXt excels in meat and gonad segmentation results. These observations underscore the impact of varying scales on the segmentation of different oyster components using different models. This variability in performance speaks to the importance of selecting a model that aligns with the specific characteristics and requirements of the images at hand.

To address the challenges associated with varying scales in oyster segmentation, our approach employs a multi-network ensemble approach that combines predictions from multiple scales strategy based on the hierarchical attention mechanism (Tao, Sapra et al. 2020). This involves combining predictions from diverse models, each contributing unique strengths and characteristics to create a comprehensive solution. This mechanism ensures that diverse models collaboratively contribute to the final segmentation outcome, addressing challenges posed by varying scales. The synergistic approach of ensemble learning, and multi-scale predictions provides a robust and versatile solution for instance segmentation across a spectrum of scenarios and image scales. Furthermore, our framework introduces local feature enhancement by incorporating each single oyster object cropped from the entire image. This augmentation refines features specific to individual oyster components, contributing to a more nuanced and detailed understanding of each oyster object.

Overall, there are two main contributions as below:

- **Integration of multi-network ensemble Learning with multi-scale predictions:** By combining predictions from various scales and utilizing the hierarchical attention mechanism, it harnesses the strengths of ensemble learning and addresses challenges posed by varying scales, ensuring a versatile and effective solution for instance segmentation across diverse components.
- **Fusing global and local scales for robust instance segmentation:** The fusing of local feature enhancement complements the multi-network ensemble and hierarchical attention mechanism, can refine the instance segmentation process for each oyster. Through these combined strategies, our framework aims to achieve comprehensive and precise instance segmentation results, ensuring a thorough understanding of oyster images at both global and local scales.

## 2. Related Work

In this section, we provide an overview of state-of-art research in three primary areas: phenotype segmentation methods, multi-network ensembles and multi-scale segmentation structures.

### 2.1 Phenotype segmentation methods in aquaculture

Phenotypic segmentation methods are currently studied primarily based on the learning visual information of image pixels by using instance segmentation algorithms (Grys et al. 2017) (Chen et al. 2018). Instance segmentation incorporates object detection learning procedure to segment all instances under consideration. Compared with semantic segmentation that focus on pixel-level learning, the instance segmentation will take into consideration with complex dynamics of individual objects get an accurate estimate of the boundary of each instance. The backbone models of the instance segmentation can be mainly focused on either the CNN-based or transformer-based methods.

A typical example of CNN-based segmentation methods is Mask-RCNN (He et al. 2017) (Garcia et al. 2020), which extended from Faster R-CNN by adding a branch for predicting an object mask in parallel with the existing branch for bounding box recognition, which has been used in many aquatic species segmentation, for example, Milena, et al. (Freitas et al. 2023) proposed research on the phenotype segmentation of body shape in pacu Piaractus mesopotamicus to facilitate the incorporation of this trait as a breeding goal. UNet (Nezla, Haridas, and Supriya 2021) is another popular CNN-based architecture for semantic segmentation, which consisted of a contracting path to capture context and a symmetric expanding path that enables precise localization. For example, UNet was introduced to segment fish body area by combining fish morphological characteristics (Yu et al. 2022). Jianyuan, et al. (Li et al. 2023) proposed algorithm (RA-UNet) that is based on ResNet50 and Unet for fish segmenting and measuring phenotypes. Some other CNN-based models, such as YOLO(Abinaya, Susan, and Sidharthan 2022) (Thayananthan et al. 2023), PSS-net (Kim and Park 2022), PSPNet and DeepLabv3 (Böer, Veeramalli, and Schramm 2021) were also presented in the recent years for marine species phenotype segmentation.

Regarding the transformed-based methods, Vision transformer (ViT) (Saleh et al. 2022) was introduced a Transformer-based method that uses self-supervision for high-quality fish body segmentation. Multi-scale transformer network (MulTNet) (Xu et al. 2022) was proposed for improving the segmentation accuracy of marine animals, and it simultaneously possessed the merits of a convolutional neural network (CNN) and a transformer. Cell detection transformer (Cell-DETR) (Hörst et al. 2023) was an attention-based detection transformer for instance segmentation. Also, by combining CNN-based and transformer-based method, Shima, et al, (Javanmardi et al. 2023) proposed a novel pipeline that integrates the Transformer and U-Net, a convolutional neural network for biomedical image segmentation, to achieve accurate segmentation of zebrafish larvae images. Mobile Fish Landmark Detection network (MFLD-net) using convolution operations based on Vision Transformers (i.e. patch embeddings, multi-layer perceptron was utilised to detect fish key points (landmarks) (Saleh et al. 2023).

However, the direct application of these models to phenotype segmentation in aquaculture faces the following problems: 1) Most phenotype segmentation requires a high-resolution phenotype datasets as training support for semantic segmentation models, however, there is a notable gap in addressing the specific challenges posed by various resolution images, which is crucial for the broader application of phenotype segmentation in diverse datasets; 2) Existing phenotype segmentation techniques utilized in aquaculture primarily concentrate on individual objects within an image. However, there is a scarcity of research on phenotype segmentation for multiple objects, like multiple oysters, within a single image. Some objects, characterized by small dimensions and intricate components, require precise segmentation; and 3). To the best of my knowledge, there is currently no research or application focused on oyster phenotype segmentation, which is crucial for advancing automatic grading process.

2.2 Multi-network ensembles

In recent years, the integration of multi-network ensembles has emerged as a powerful strategy in the field of computer vision. These approaches leverage the strengths of multiple neural networks to enhance the robustness and overall performance.

The utilization of ensembles, incorporating diverse neural networks, has demonstrated notable advantages in handling complex images in many domains, such as medical images (Mahbod et al. 2020) (Wang et al. 2020) (Zhang and Gao 2021) (Sakib and Siddiqui 2023), biological images (Peng et al. 2021) (Li et al. 2022) (Mufassirin et al. 2023) and agriculture images (Amudha and Brindha 2022) (Velásquez, Lara, and Velásquez 2023). By combining predictions from multiple networks, the ensemble approach mitigates individual network limitations, resulting in improved accuracy and generalization. Noteworthy techniques include bagging, boosting, and stacking (Odegua 2019), each contributing to the ensemble's ability to capture intricate features within images.

However, multi-network ensembles still have limited research and application in aquaculture, with a focus on ensembled machine learning techniques. For instance, there have been studies developing ensembled machine learning models, such as linear regression-based prediction models for marine fish and aquaculture production (Rahman et al. 2021), predicting for disease resistance in aquaculture species using machine learning models (Palaiokostas 2021) and predicting growth of abalone reared in land-based aquaculture using machine learning ensembles (Khiem et al. 2023). In the context of phenotype segmentation in oysters, there is a need to develop more complex multi-model ensembles for segmentation. These ensembles should be designed to handle diverse components, conditions, and populations, aiming to create a comprehensive understanding of the phenotypic variations.

## 2.3 Multi-scale segmentation structures in segmentation

Addressing the challenges posed by varying resolutions in image segmentation, multi-scale methods have gained prominence. These methods operate across different image scales, allowing for the effective segmentation of objects with diverse sizes. This approach was first proposed by Chen et al. to capture multi-scale context features within the framework of FCN for dense pixel prediction (Chen et al. 2016). Subsequently, several variants based on different attention mechanisms have been proposed, including self-guided attention (Sinha and Dolz 2020), dual attention (Wang, Wang, et al. 2021), hierarchical multi-scale attention (Tao, Sapra, and Catanzaro 2020) (Heidari et al. 2023), cascade attention (Rahman and Marculescu 2024), all contributing to advancements in semantic segmentation. Table 2 shows the comparison analysis between these different attention mechanisms.

Table 2. Comparison analysis between these different attention mechanisms in semantic segmentation

|  | Key features | Limitations | Typical applicable images | References |
|---|---|---|---|---|
| Explicit attention | learn all attention masks for each of a fixed | May not adapt well to variable scales or sizes. | Images with static scales | (Chen et al. 2016) (Wang, Zhang, et al. 2021) (Hu et al. 2021) |
| | set of scales. | | | |
| Self-guided attention | add progressive refinement of attentive features through sequential refinement modules | May introduce computational overhead due to sequential processing. | Images with complex structures or diverse content | (Sinha and Dolz 2020) (Xu et al. 2021) (Karimijafarbigloo et al. 2024) (Usman et al. 2024) |
| Dual attention | Add semantic correlation by combining the channel and spatial mechanism | Use the attention module multiple times will increase the memory burden and consume more resources | Images with significant emphasis on spatial features, such as scene understanding. | (Wang, Wang, et al. 2021) (Guo et al. 2021) (Ji et al. 2023) |
| Hierarchical attention | Learn relative attention masks between adjacent scales. | Learning relative attention masks between different scales might introduce additional computational complexity, particularly during the training phase. | Images with diverse scales and different segmentation levels. | (Tao, Sapra, and Catanzaro 2020) (Heidari et al. 2023) (Sun, Shao, et al. 2023) |
| Cascade attention | Aggregate multi-scale features through a sequential cascade. | May have limitations in capturing long-range dependencies or global context, as each stage focuses on a local region | Images with rich and intricate details | (Sun, Dai, et al. 2023) (Zhou et al. 2023) (Rahman and Marculescu 2024) |

Based on the above analysis, concerning the segmentation of oyster phenotype images, which are captured using various methods resulting in diverse scales and resolutions, it is noted that oyster phenotype segmentation comprises different levels of segmentation corresponding to distinct components. However, spatial features and rich details may not be critical considerations in this context. Furthermore, each image may encompass multiple objects, necessitating the capture of both global features (indicating the types of components across all oysters) and local features (specific to each oyster and its components). These observations suggest that hierarchical attention may be well-suited for the task of oyster phenotype segmentation.

## 3. Method

In our study, we proposed a multi-Network ensemble and multi-scale mechanism for Pacific oyster phenotype segmentation by using different backbones with the ensemble method. Our network architecture is an extension of the hierarchical multi-scale attention model proposed by Tao et al (Tao, Sapra, and Catanzaro 2020). In our adaptation, we integrate this attention mechanism with a multi-network ensemble and incorporate a local attention component. This innovative combination allows our proposed network architecture to leverage the strengths of different models, providing a robust framework that considers both global and local context for improved segmentation accuracy. Fig. 2 and Fig. 3 shows the overall network architecture for training and inference respectively.

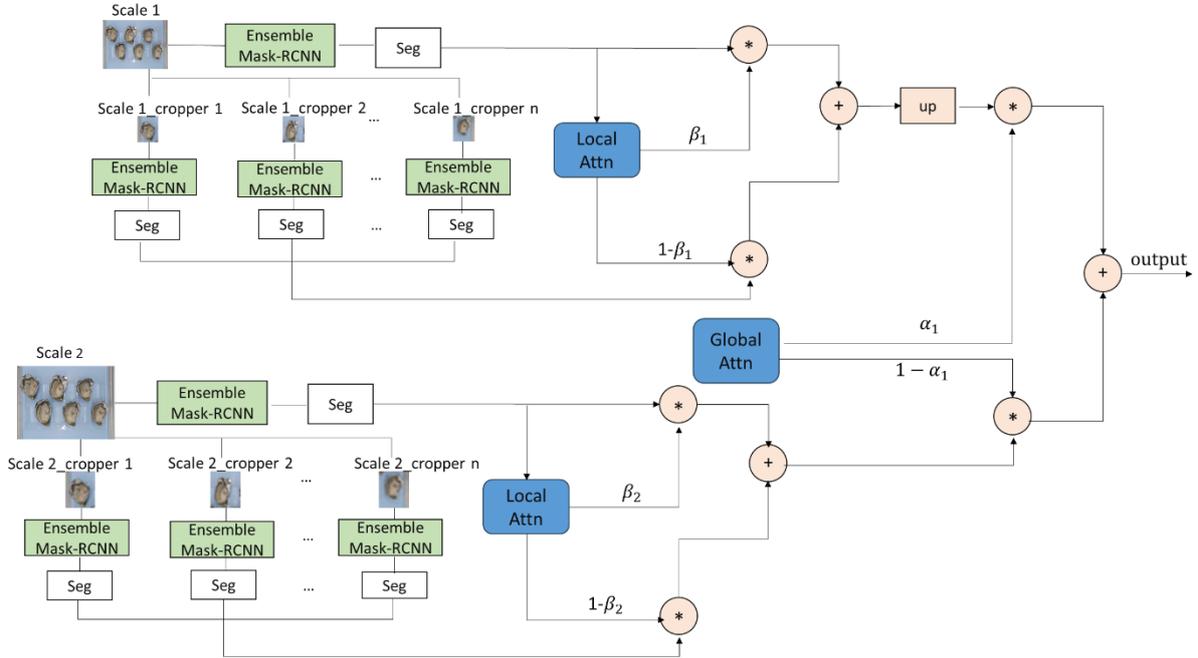

Fig.2. Training network: learning across adjacent scale pairs

During training, we utilized adjacent scale pairs (e.g., scales *m* and *n*, where *j* is one level upscale for *m* among multiple scales). For each scale, the multi-network ensemble is employed to produce the global semantic logit $L_{global}^{m}$ and each local semantic logit $L_{local}^{p}$, where *q* represents the number of oysters in the entire image. Subsequently, the local attention $\beta_m$ for scale *m* is added to fuse the global and local semantic logits.

In the next step, to combine the semantic logits from two adjacent scales, the global attention $\alpha_i$ is introduced along with an up-sampling operation. This operation involves pixel-wise multiplication and addition to obtain the final output. The overall formalization of these two steps is represented by equations (1) and (2) below:

$$L^i = E\left(L_{global}^m\right)\odot\beta_m + \sum_{p=1}^{q} E\left(L_{local}^p\right)\odot(1-\beta_m) \quad (1)$$

$$L_{output} = U(L^m) \odot \alpha_m + L \odot (1 - \alpha_m) \tag{2}$$

Here, *E* signifies the multi-network ensembles, *U* signifies the up-sampling operation, and $\odot$ represents pixel-wise multiplication.

For the global attention weights $\alpha_m$, we utilised the attention mechanism inspired by (Tao, Sapra, and Catanzaro 2020) to weigh the information from different scales. The generation of local attention weights will be introduced in Section 3.3.

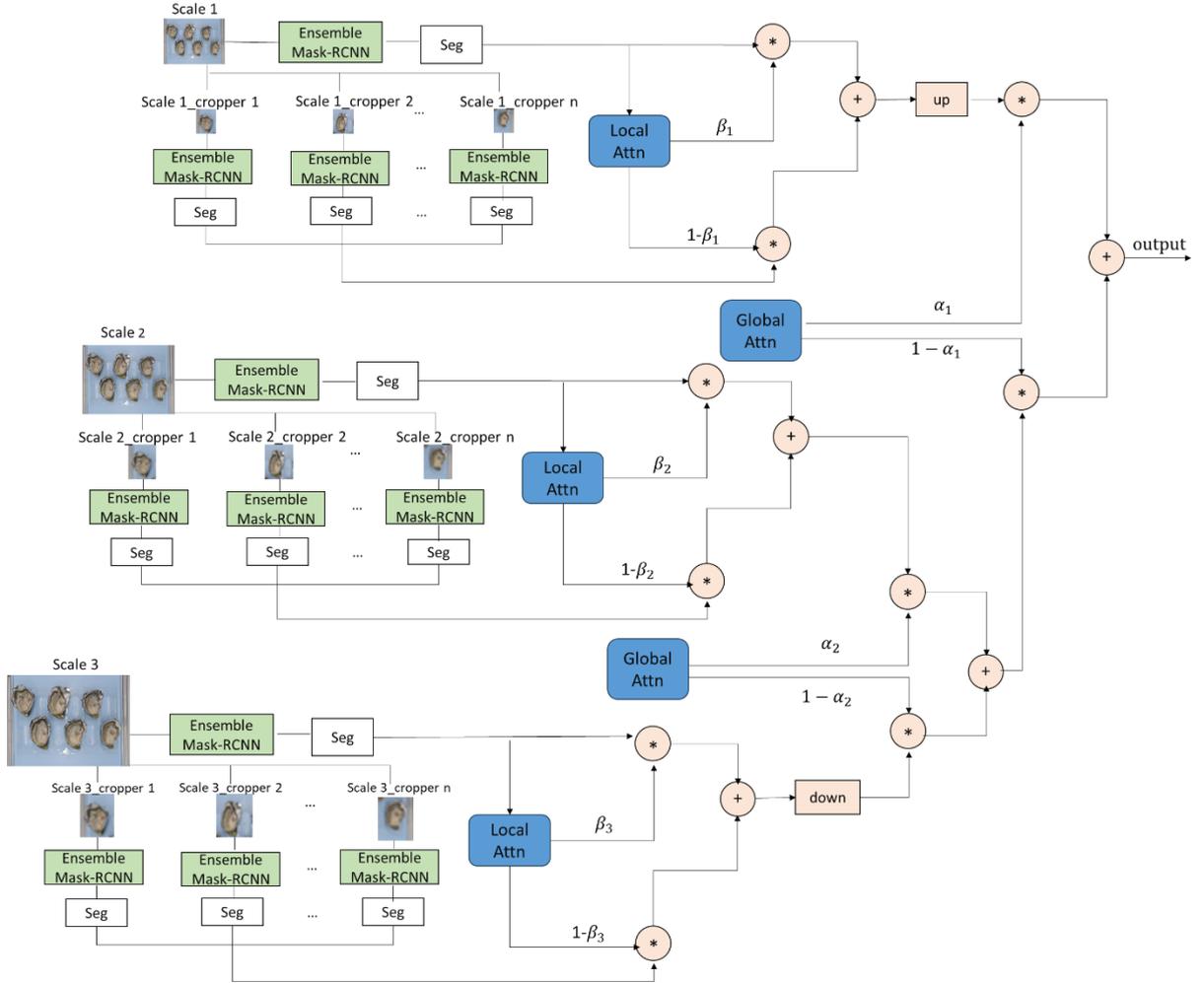

Fig.3. Inference network: hierarchical fusion of multiple scales

Regarding the process of inference, which is strategically designed to amalgamate predictions across multiple scales. The lower-scale attention not only shapes the precision of predictions at its own level but, significantly, acts as the guiding force that determines the influence and contribution of the subsequent higher scale. The interplay of attention across different scales forms a seamless continuum, creating a cohesive and interlinked framework for generating refined and multi-scale predictions.

## 3.1 Datasets

In our research, we strategically centre our attention on small datasets, recognising the commonality of compact private datasets in the field of aquaculture. Our approach to image collection involves placing multiple Pacific oysters on trays for setting different capture conditions. We carefully account for various settings that lead to variations in resolution and size, owing to the utilization of different devices. The detailed breakdown in Table 3 provides a comprehensive overview of the three specific datasets we've meticulously collected. In total, our study comprises three private datasets, totalling 171 images. These small datasets serve to demonstrate the feasibility and efficiency of our proposed methods in handling such limited-sized datasets.

Table 3. Overview of three Collected Oyster Datasets

|  | Dataset 1 | Dataset 2 | Dataset 3 |
|---|---|---|---|
| Device | GoPro 9 | GoPro 9 | Cannon |
| Resolution | 72 dpi | 24 dpi | 180 dpi |
| Size | 3504*2624 | 3504*2624 | 5152*3864 |
| Train # | 43 | 31 | 43 |
| Validate # | 10 | 7 | 10 |
| Test # | 10 | 7 | 10 |
| Total # | 63 | 45 | 63 |
| Typical example | 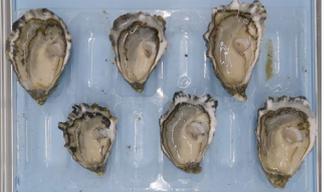 | 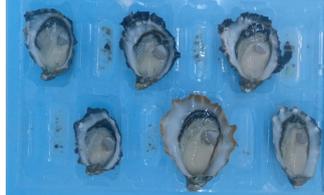 | 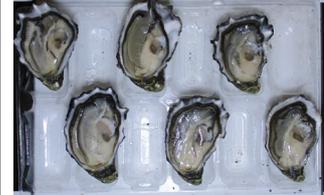 |

For cropping, each oyster is extracted from the entire image based on the detected bounding box (IoU=0.6) shown in Fig.4., To ensure that the entire shell is encompassed in the cropped images, we upscale the size of the bounding box to 1.2. In cases where certain bounding boxes are mis-detected or incomplete, additional post-processing techniques, such as manual refinement or algorithmic correction, may be applied to enhance the accuracy of the bounding box representations. These measures contribute to maintaining the overall integrity of the cropped oyster images, even in instances of imperfect initial detection.

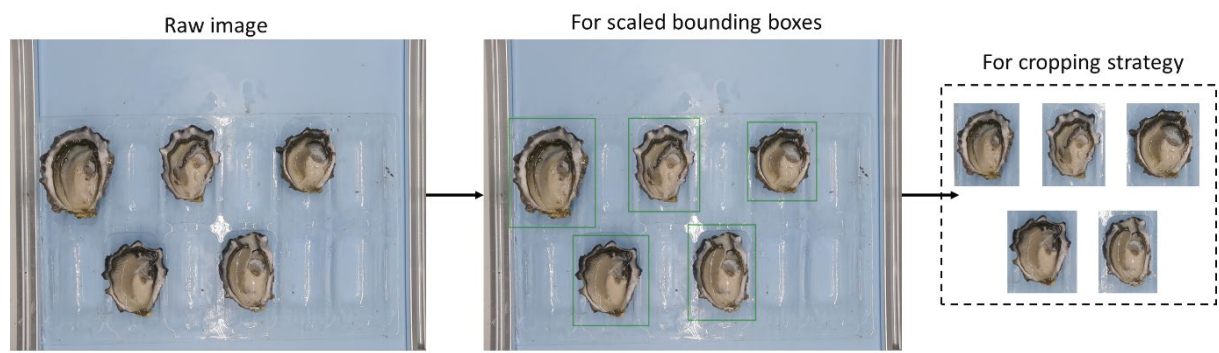

Fig.4. Examples of the cropping strategy based on the raw images.

## 3.2 Ensemble Mask-RCNN

To build the ensemble Mask-RCNN, we establish the Weighted Segmentation Mask Fusion. We have segmentation masks predictions for the same image containing M oysters from N different models, so total we will get M*N*4 predictions. We utilized the ensemble Mask-RCNN for both vertical (for each component) and horizontal (for each oyster) applications. The overall architecture of the ensemble Mask-RCNN is illustrated in Fig. 5, functioning through the following steps.

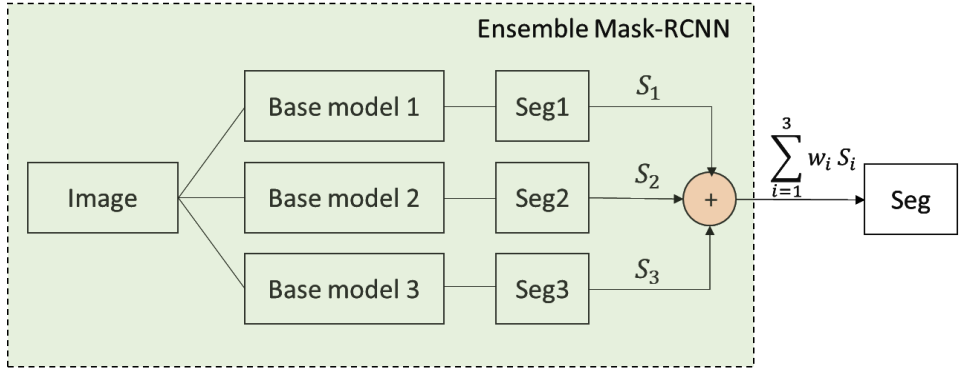

Fig.5. The weighted segmentation mask fusion strategy

**Aggregation and group of predicted seg masks:** Each predicted segmentation mask from every model is gathered into a single list, denoted as S. This list is then sorted in descending order based on the segmentation precision (P) associated with each mask. Additionally, segmentation objects within the list can be grouped based on components (vertical grouping), and alternatively, they can be grouped based on oysters (horizontal grouping).

**Weight Calculations:** For the vertical weights pertaining to the entire image, we will compute the segmentation AP (Average Precision) for each model with respect to specific components across all oysters. Similarly, for the horizontal weights concerning the cropped images, we will calculate the segmentation AP for each model with respect to each oyster among all the components. Subsequently, normalize the segmentation AP scores to a scale between 0 and 1. This ensures that models with varying AP scales can be compared uniformly. Assign weights to each model based on its normalized segmentation AP score, where higher AP scores should result in higher weights.

**Fused Mask Generation:** calculate the weighted average of the corresponding segmentation masks. Let $S_{fused}$ represent the fused mask, $S_i$ represent the masks from different models in the same group and $w_i$ represent the weight for each model based on different group strategies. Normalized $AP_i$ is the normalised segmentation AP score for Model i, j is the number of models used in the ensembles.

$$S_{fused} = \sum_i \frac{Normalized\ AP_i}{\sum_j Normalized\ AP_j} S_i \qquad (3)$$

This approach ensures that fused masks are organized by both components and oysters inherent in your segmentation tasks. By incorporating information from different models and considering specific components and oysters, the fusion process aims to enhance the overall segmentation accuracy.

### 3.3 Global- local hierarchical attention

In our study, we propose a global-local hierarchical attention mechanism to enhance the capability of considering context. To generate local attention weights for each image in our segmentation task, we leverage the RoIAlign layer of the ensemble Mask R-CNN to extract both local RoI features from cropped images and global RoI features from the entire image. Instead of directly concatenating these local and global features, we employ a difference-based attention mechanism to combine the global RoI feature map and local RoI feature map. The overall local attention implementation is shown in Fig. 6.

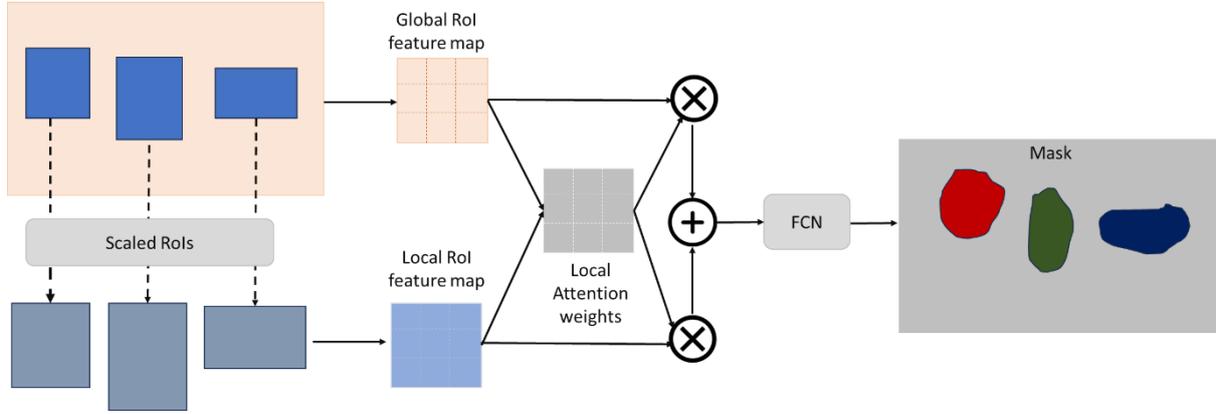

**Fig.6.** The local attention mechanism implementation

In detail, the following steps are outlined below.

First, we define the difference matrix *D*: Given the global feature map *G* and local feature map *L*, calculate the absolute difference between corresponding elements.

$$D_{ij} = |G_{ij} - L_{ij}| \quad (4)$$

Secondly, the attention matrix $\beta$ is typically obtained using the softmax function applied to the negative absolute differences, often scaled by a factor *f* for better numerical stability. The softmax operation is applied row-wise for each element in *D*. The softmax function ensures that the attention weights are in the range [0, 1] and represent the relative importance of each element in the feature maps. It effectively converts the differences into a probability distribution.

$$\beta_{ij} = \frac{e^{-f|D_{ij}|}}{\sum_k e^{-f|D_{ik}|}} \quad (5)$$

Here *k* represents the index variable for the summation across all elements in the same row.

Finally, we normalize the attention weights across each row (local features). The normalization step ensures that the attention weights are distributed proportionally within each row, representing the contribution of each local feature to the global context.

$$\beta_{ij} = \frac{A_{ij}}{\sum_j A_{ij}} \quad (6)$$

Based on the calculated local attention weights, we subsequently incorporate hierarchical global attention to generate the final outputs using equations (1) and (2).

**4. Experimental results and analysis**

To evaluate the effectiveness of our proposed approach, we trained networks using three backbones: ResNet50-FPN, R101-FPN, and ResNeXt-FPN, along with four datasets collected as listed in Table 1. All experiments were conducted on a single workstation equipped with an Intel(R) Core (TM) i7-1365U 1.80 GHz CPU, 32GB of RAM, and an Nvidia Tesla V100 GPU.

4.1 Comparison of ensemble method

To assess the effectiveness of the proposed ensemble Mask-RCNN, we conducted experiments utilizing various base backbones, including R50-FPN, R101-FPN, and ResNeXt-FPN, across different scales (0.25, 0.5, 1.0). The evaluation encompassed both single-image scenarios by combining dataset 1 and dataset 2. We compared the Mask Average Precision (Mask AP) for different components (S represents shell, B represents the body meat, G represents the gonad, and M represents the muscle) for individual networks against our proposed multi-network ensemble approach.

Table 4. Comparison of our ensemble network model vs single models for different scaled images

|  |  | R50-FPN | R101-FPN | ResNeXt-FPN | Multi-network Ensemble (ours) |
|---|---|---|---|---|---|
| Scale1: 1.0 | Shell | 91.19 | 91.76 | 91.79 | 93.51 |
|  | Meat | 69.50 | 61.18 | 73.69 | 79.85 |
|  | Gonad | 42.42 | 43.65 | 58.83 | 57.98 |
|  | Muscle | 37.68 | 35.63 | 41.20 | 48.05 |
| Scale 2: 0.5 | Shell | 90.06 | 90.41 | 90.33 | 92.06 |
|  | Meat | 70.18 | 63.48 | 69.17 | 76.97 |
|  | Gonad | 41.23 | 56.00 | 49.72 | 57.98 |
|  | Muscle | 30.45 | 32.80 | 37.79 | 45.54 |
| Scale 3: 2.0 | Shell | 89.28 | 92.33 | 90.62 | 93.57 |
|  | Meat | 66.62 | 46.33 | 73.12 | 74.86 |
|  | Gonad | 55.27 | 34.67 | 60.59 | 60.30 |
|  | Muscle | 35.96 | 36.37 | 42.42 | 47.58 |

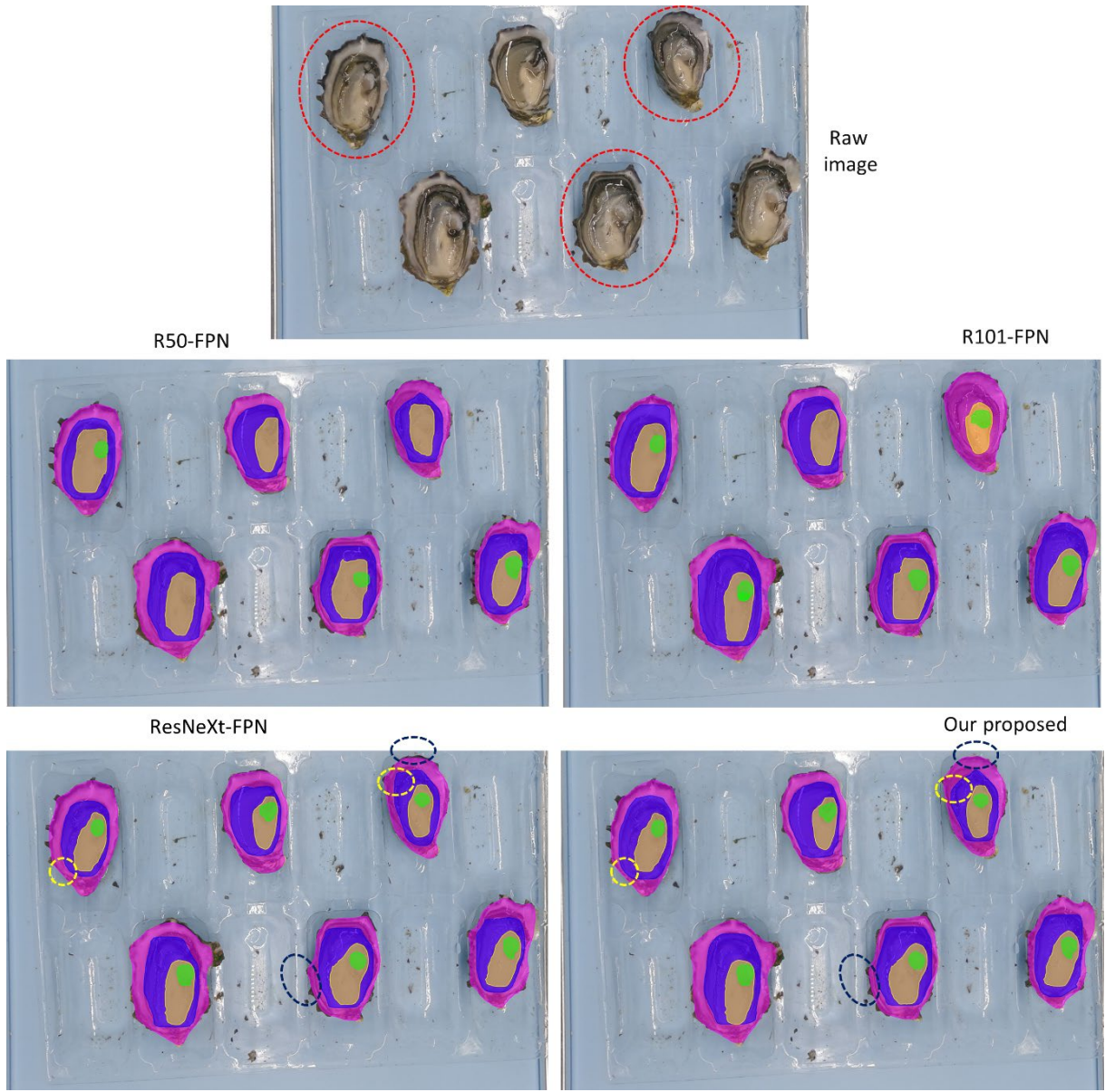

**Fig.7.** Sample segmentation results (use scale 1.0) by comparing single model and the proposed ensemble model.

As shown in Table 4, the proposed multi-network ensemble consistently outperforms individual models across all image scales and categories. When examining scales 1.0 and 2.0, the accuracy percentages for Shell, Meat, and Muscle exhibit substantial enhancements, while the Gonad category demonstrates a maintained accuracy level. The most striking advantage of our proposed ensemble approach becomes evident at Scale 0.5. In this lower-resolution scenario, where challenges in image recognition often arise, our model excels with remarkable accuracy improvements, showcasing its robustness in handling lower-resolution images. Additionally, the sample segmentation results depicted in Fig. 7 represent a comparison between a single model and our proposed ensemble model. The visual comparison clearly illustrates that our proposed method captures finer details, such as precise boundary delineation for the shell

(examples outlined by blue dashed lines) and enhanced contours that highlight details in the meat (examples delineated by yellow dashed lines).

4.2 Comparison of Global-Local Hierarchical Attention among different image scales

Section 4.1 has been approved for its effectiveness through the use of the multi-network ensemble method. Building upon the proposed ensemble method, this section aims to assess efficiency by incorporating local attention to enable global-local hierarchical attention. Within this section, we utilized datasets 1, 2, and 3 as outlined in Table 5 to compare segmentation Average Precision (AP) across various image scales.

Table 5. Comparison results across various scales and attention mechanisms.

| Dataset | Methods | Eva scale | Segm AP | | | |
|---|---|---|---|---|---|---|
| | | | Shell | Meat | Gonad | Muscle |
| Dataset 1 | Single scale without local attention | 1.0 | 92.51 | 79.17 | 46.64 | 53.25 |
| | Single scale with local attention | 1.0 | 97.50 | 78.56 | 51.41 | 60.31 |
| | Single scale without local attention | 0.5 | 90.92 | 79.21 | 52.78 | 51.17 |
| | Single scale with local attention | 0.5 | 97.12 | 79.13 | 57.81 | 60.94 |
| | Multi-scale without local attention | 0.25, 0.5, 1.0 | 92.50 | 79.79 | 53.11 | 53.71 |
| | Multi-scale with local attention | 0.25, 0.5, 1.0 | 97.71 | 80.38 | 57.62 | 60.91 |
| | Multi-scale without local attention | 0.5, 1.0, 2.0 | 92.79 | 79.75 | 53.85 | 54.01 |
| | Multi-scale with local attention | 0.5, 1.0, 2.0 | 97.56 | 79.84 | 57.68 | 60.67 |
| Dataset 2 | Single scale without local attention | 1.0 | 95.68 | 81.81 | 67.83 | 46.68 |
| | Single scale with local attention | 1.0 | 98.74 | 84.17 | 77.08 | 62.75 |
| | Single scale without local attention | 0.5 | 94.14 | 75.94 | 64.43 | 46.05 |
| | Single scale with local attention | 0.5 | 98.67 | 84.00 | 72.67 | 61.20 |
| | Multi-scale without local attention | 0.25, 0.5, 1.0 | 96.04 | 81.84 | 67.94 | 47.18 |
| | Multi-scale with local attention | 0.25, 0.5, 1.0 | 99.13 | 85.00 | 77.25 | 63.65 |
| | Multi-scale without local attention | 0.5, 1.0, 2.0 | 96.01 | 81.87 | 67.48 | 47.43 |
| | Multi-scale with local attention | 0.5, 1.0, 2.0 | 98.71 | 84.07 | 77.44 | 63.51 |
| Dataset 3 | Single scale without local attention | 1.0 | 86.70 | 84.47 | 65.73 | 47.61 |
| | Single scale with local attention | 1.0 | 94.54 | 87.27 | 66.36 | 57.27 |

| | Single scale without local attention | 0.5 | 85.56 | 84.39 | 67.82 | 44.04 |
| --- | --- | --- | --- | --- | --- | --- |
| | Single scale with local attention | 0.5 | 94.06 | 87.81 | 69.06 | 54.38 |
| | Multi-scale without local attention | 0.25, 0.5, 1.0 | 87.16 | 84.98 | 67.49 | 48.40 |
| | Multi-scale with local attention | 0.25, 0.5, 1.0 | 94.70 | 87.82 | 69.27 | 58.24 |
| | Multi-scale without local attention | 0.5, 1.0, 2.0 | 87.63 | 84.52 | 67.13 | 48.05 |
| | Multi-scale with local attention | 0.5, 1.0, 2.0 | 95.12 | 88.05 | 69.33 | 59.24 |

**Effect of Attention Mechanism:** Effect of Attention Mechanism: When comparing methods with and without local attention, we consistently observe improvements across all individual scales and segmentation average precision (Segm AP) for various components. Specifically, the average improvements of SegmAP for shell, meat, gonad, and muscle are 6.45%, 3.31%, 7.89%, and 23.63% respectively. Fig.8. shows the Segm AP performance comparison between models without and with local attention. Based on the significant improvement observed, particularly in the muscle component, it suggests that the addition of local attention introduces more detailed features, leading to substantial performance gains.

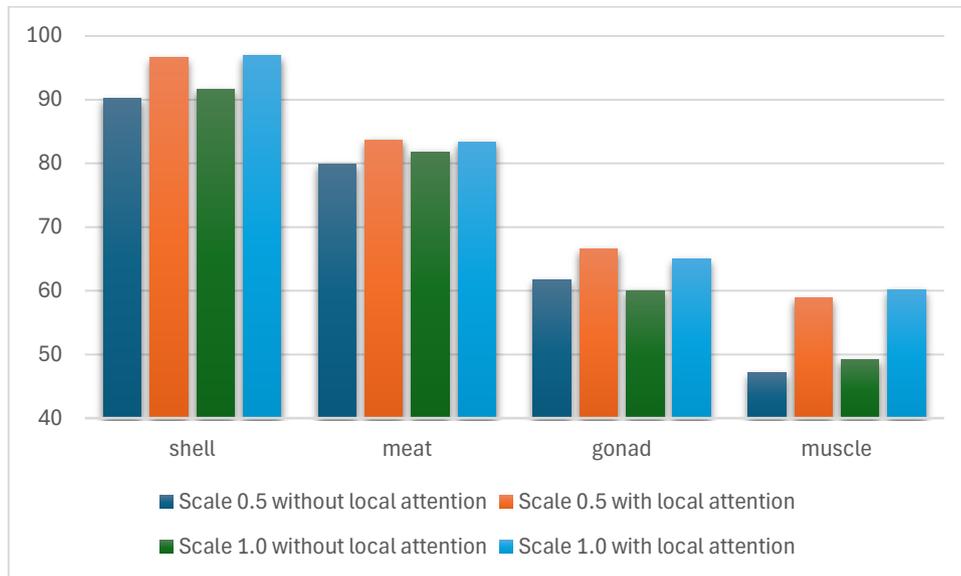

**Fig.8.** The performance comparison between models without and with local attention

**Effect of Scale Variation:** When examining the impact of scale variation, we calculate the average improvements by comparing two multi-scale configurations, [0.25, 0.5, 1.0] and [0.5, 1.0, 2.0], with single scales of 1.0 and 0.5, both employing local attention. Across all datasets, the implementation of multi-scale approaches consistently results in improvements in the performance of shell, meat, gonad, and muscle components compared to single-scale approaches. Specifically, the gonad and muscle components show the highest average improvements of 2.37% and 1.61%, respectively, across all three datasets. Fig. 9 illustrates the performance improvements achieved by using multi-scales [0.25, 0.5, 1.0] and [0.5, 1.0, 2.0], compared to using the relevant single scale only, across three different datasets.

Lastly, Figure 10 displays the phenotype segmentation of sample oyster images extracted from three distinct datasets.

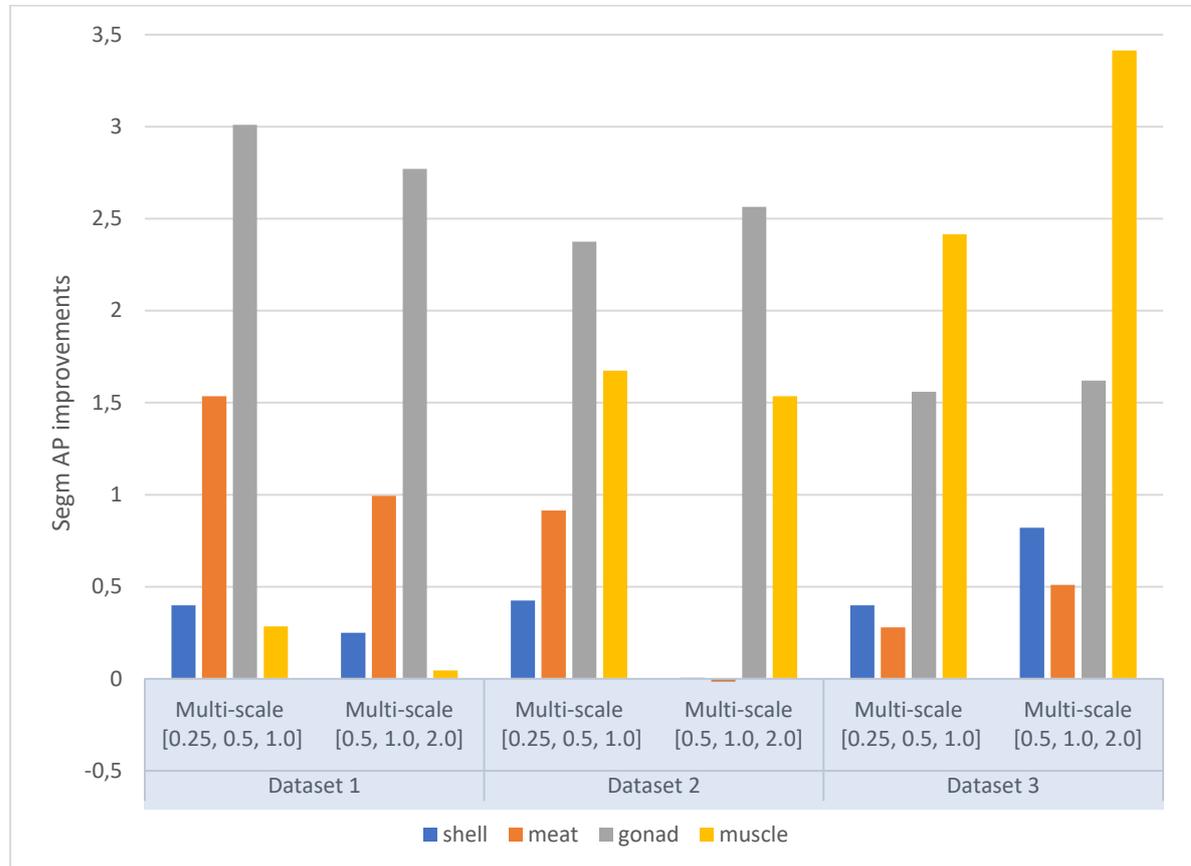

**Fig.9.** The performance comparison between models using multi-scales and single scales with local attention.

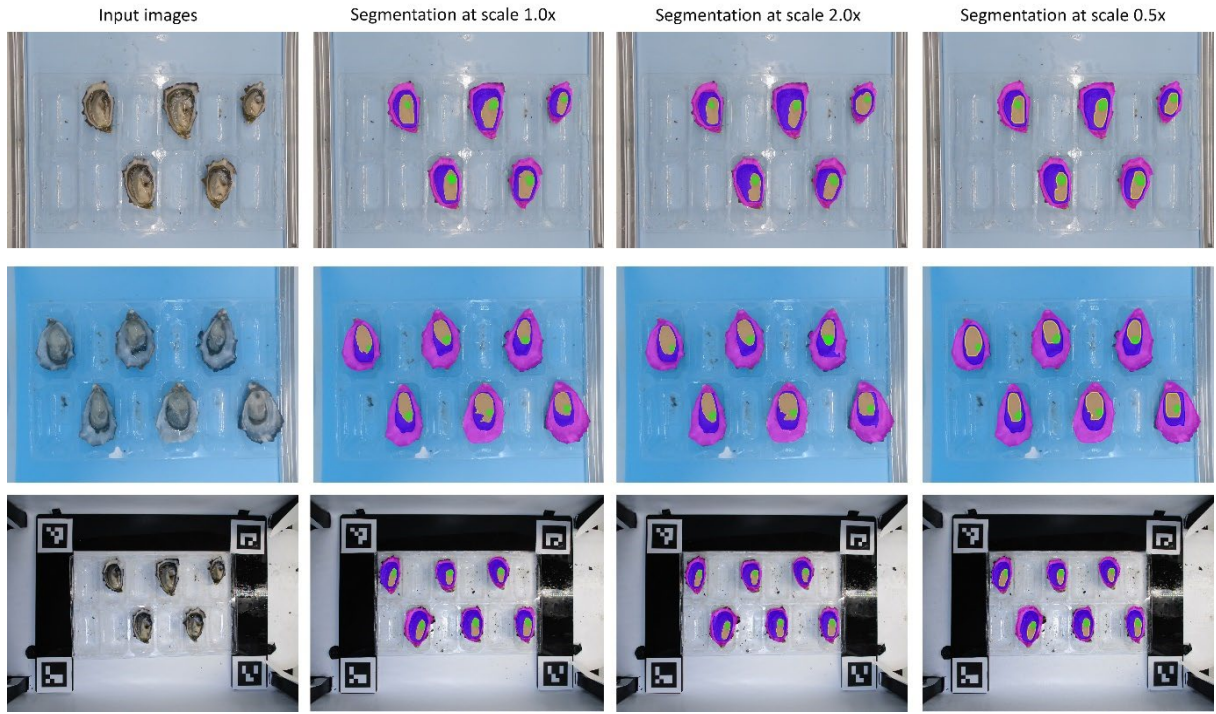

**Fig. 10.** Sample phenotype segmentation results among different datasets using our proposed method.

## 5. Conclusion

Our study highlights the significance of phenotype segmentation and presents a novel multi-network ensemble approach with a global-local hierarchical attention mechanism. In our proposed method, we introduce two key innovations. Firstly, we integrate multi-network ensemble learning with multi-scale predictions, leveraging diverse model predictions and employing a hierarchical attention mechanism to address challenges posed by varying scales. This ensures a versatile and effective solution for instance segmentation across diverse oyster components. Secondly, we propose a method for fusing global and local features into the attention mechanism. By enhancing local features and combining them with the multi-network ensemble and hierarchical attention mechanism, our framework aims to refine the instance segmentation process for each component (shell, meat, gonad, and muscle), resulting in comprehensive and precise segmentation results. These contributions significantly advance the field of oyster phenotype segmentation, facilitating a deeper understanding of oyster images.

Looking ahead, there are several avenues for future research. Firstly, extending the testing of our approach to more diverse datasets would further validate its performance across various environments. Additionally, incorporating additional phenotype characteristics such as shape, size, and 3D features could provide a more comprehensive understanding of oyster traits. Moreover, integrating our segmentation approach into meat condition ranking systems has the potential to enhance overall quality assessment processes and contribute to advancements in oyster production and management practices. These future endeavours hold promise for further advancing the field of oyster phenotype segmentation and its applications in aquaculture and related industries.

**Funding and Acknowledgement**

This work was supported by the FRDC (project number: 2021-083). Henry Hewish, Lewa Pertl, Matthew Cunningham and Nick Griggs of Australian Seafood Industries are thanked for their input in the coordination of the FRDC project. The Sustainable Marine Research Collaboration Agreement (SMRCA) between the Department of Natural Resources and Environment, Tasmania (DNRE), and the University of Tasmania is acknowledged for supporting oyster aquaculture research at IMAS.

**References**

Abinaya, NS, D Susan, and Rakesh Kumar Sidharthan. 2022. 'Deep learning-based segmental analysis of fish for biomass estimation in an occulted environment', *Computers and Electronics in Agriculture*, 197: 106985.

Amudha, M, and K Brindha. 2022. 'Multi Techniques for Agricultural Image Disease Classification and Detection: A Review', *Nature Environment & Pollution Technology*, 21.

Böer, Gordon, Rajesh Veeramalli, and Hauke Schramm. 2021. "Segmentation of Fish in Realistic Underwater Scenes using Lightweight Deep Learning Models." In *ROBOVIS*, 158-64.

Cao, Jiaheng, and Lihong Xu. 2018. "Research on counting algorithm of residual feeds in aquaculture based on machine vision." In *2018 IEEE 3rd International Conference on Image, Vision and Computing (ICIVC)*, 498-503. IEEE.

Chen, Liang-Chieh, Yi Yang, Jiang Wang, Wei Xu, and Alan L Yuille. 2016. "Attention to scale: Scale-aware semantic image segmentation." In *Proceedings of the IEEE conference on computer vision and pattern recognition*, 3640-49.

Chen, Weiyang, Weiwei Li, Xiangjun Dong, and Jialun Pei. 2018. 'A review of biological image analysis', *Current Bioinformatics*, 13: 337-43.

Daud, Ahmad Kamal Pasha Mohd, Norakmar Arbain Sulaiman, Yuslinda Wati Mohamad Yusof, and Murizah Kassim. 2020. "An IoT-based smart aquarium monitoring system." In *2020 IEEE 10th Symposium on Computer Applications & Industrial Electronics (ISCAIE)*, 277-82. IEEE.

Freitas, Milena V, Celma G Lemos, Raquel B Ariede, John FG Agudelo, Rubens RO Neto, Carolina HS Borges, Vito A Mastrochirico-Filho, Fábio Porto-Foresti, Rogério L Iope, and Fabrício M Batista. 2023. 'High-throughput phenotyping by deep learning to include body shape in the breeding program of pacu (Piaractus mesopotamicus)', *Aquaculture*, 562: 738847.

Garcia, Rafael, Ricard Prados, Josep Quintana, Alexander Tempelaar, Nuno Gracias, Shale Rosen, Håvard Vågstøl, and Kristoffer Løvall. 2020. 'Automatic segmentation of fish using deep learning with application to fish size measurement', *ICES Journal of Marine Science*, 77: 1354-66.

Grys, Ben T, Dara S Lo, Nil Sahin, Oren Z Kraus, Quaid Morris, Charles Boone, and Brenda J Andrews. 2017. 'Machine learning and computer vision approaches for phenotypic profiling', *Journal of Cell Biology*, 216: 65-71.

Guo, Pengcheng, Xiangdong Su, Haoran Zhang, and Feilong Bao. 2021. "Mcdalnet: Multi-scale contextual dual attention learning network for medical image segmentation." In *2021 International Joint Conference on Neural Networks (IJCNN)*, 1-8. IEEE.

He, Kaiming, Georgia Gkioxari, Piotr Dollár, and Ross Girshick. 2017. "Mask r-cnn." In *Proceedings of the IEEE international conference on computer vision*, 2961-69.

He, Kaiming, Xiangyu Zhang, Shaoqing Ren, and Jian Sun. 2016. "Deep residual learning for image recognition." In *Proceedings of the IEEE conference on computer vision and pattern recognition*, 770-78.

Heidari, Moein, Amirhossein Kazerouni, Milad Soltany, Reza Azad, Ehsan Khodapanah Aghdam, Julien Cohen-Adad, and Dorit Merhof. 2023. "Hiformer: Hierarchical multi-scale representations using transformers for medical image segmentation." In *Proceedings of the IEEE/CVF Winter Conference on Applications of Computer Vision*, 6202-12.

Hörst, Fabian, Moritz Rempe, Lukas Heine, Constantin Seibold, Julius Keyl, Giulia Baldini, Selma Ugurel, Jens Siveke, Barbara Grünwald, and Jan Egger. 2023. 'Cellvit: Vision transformers for precise cell segmentation and classification', *arXiv preprint arXiv:2306.15350*.

Hu, Jingfei, Hua Wang, Jie Wang, Yunqi Wang, Fang He, and Jicong Zhang. 2021. 'SA-Net: A scale-attention network for medical image segmentation', *PloS one*, 16: e0247388.

Javanmardi, Shima, Xiaoqin Tang, Mehrdad Jahanbanifard, and Fons J Verbeek. 2023. "Unsupervised Segmentation of High-Throughput Zebrafish Images Using Deep Neural Networks and Transformers." In *International Conference on Data Science and Artificial Intelligence*, 213-27. Springer.

Ji, Qiulang, Jihong Wang, Caifu Ding, Yuhang Wang, Wen Zhou, Zijie Liu, and Chen Yang. 2023. 'DMAGNet: Dual‐path multi‐scale attention guided network for medical image segmentation', *IET Image Processing*, 17: 3631-44.

Karimijafarbigloo, Sanaz, Reza Azad, Amirhossein Kazerouni, and Dorit Merhof. 2024. "Ms-former: Multi-scale self-guided transformer for medical image segmentation." In *Medical Imaging with Deep Learning*, 680-94. PMLR.

Khiem, Nguyen Minh, Yuki Takahashi, Tomohiro Masumura, Genki Kotake, Hiroki Yasuma, and Nobuo Kimura. 2023. 'A machine learning ensemble approach for predicting growth of abalone reared in land-based aquaculture in Hokkaido, Japan', *Aquacultural Engineering*, 103: 102372.

Kim, Yu Hwan, and Kang Ryoung Park. 2022. 'PSS-net: Parallel semantic segmentation network for detecting marine animals in underwater scene', *Frontiers in Marine Science*, 9: 1003568.

Li, Haozheng, Yihe Pang, Bin Liu, and Liang Yu. 2022. 'MoRF-FUNCpred: molecular recognition feature function prediction based on multi-label learning and ensemble learning', *Frontiers in Pharmacology*, 13: 856417.

Li, Jianyuan, Chunna Liu, Zuobin Yang, Xiaochun Lu, and Bilang Wu. 2023. 'RA-UNet: an intelligent fish phenotype segmentation method based on ResNet50 and atrous spatial pyramid pooling', *Frontiers in Environmental Science*.

Mahbod, Amirreza, Gerald Schaefer, Chunliang Wang, Georg Dorffner, Rupert Ecker, and Isabella Ellinger. 2020. 'Transfer learning using a multi-scale and multi-network ensemble for skin lesion classification', *Computer methods and programs in biomedicine*, 193: 105475.

Mufassirin, MM Mohamed, MA Hakim Newton, Julia Rahman, and Abdul Sattar. 2023. 'Multi-S3P: Protein Secondary Structure Prediction with Specialized Multi-Network and Self-Attention-based Deep Learning Model', *IEEE Access*.

Nezla, NA, TP Mithun Haridas, and MH Supriya. 2021. "Semantic segmentation of underwater images using unet architecture based deep convolutional encoder decoder model." In *2021 7th International Conference on Advanced Computing and Communication Systems (ICACCS)*, 28-33. IEEE.

Odegua, Rising. 2019. "An empirical study of ensemble techniques (bagging, boosting and stacking)." In *Proc. Conf.: Deep Learn. IndabaXAt*.

Palaiokostas, Christos. 2021. 'Predicting for disease resistance in aquaculture species using machine learning models', *Aquaculture Reports*, 20: 100660.

Peng, Jiajie, Hansheng Xue, Zhongyu Wei, Idil Tuncali, Jianye Hao, and Xuequn Shang. 2021. 'Integrating multi-network topology for gene function prediction using deep neural networks', *Briefings in bioinformatics*, 22: 2096-105.

Rahman, Labonnah Farzana, Mohammad Marufuzzaman, Lubna Alam, Md Azizul Bari, Ussif Rashid Sumaila, and Lariyah Mohd Sidek. 2021. 'Developing an ensembled machine learning prediction model for marine fish and aquaculture production', *Sustainability*, 13: 9124.

Rahman, Md Mostafijur, and Radu Marculescu. 2024. "Multi-scale hierarchical vision transformer with cascaded attention decoding for medical image segmentation." In *Medical Imaging with Deep Learning*, 1526-44. PMLR.

Sakib, Mohd, and Tamanna Siddiqui. 2023. 'Multi-Network-Based Ensemble Deep Learning Model to Forecast Ross River Virus Outbreak in Australia', *International Journal of Pattern Recognition and Artificial Intelligence*, 37: 2352015.

Saleh, Alzayat, David Jones, Dean Jerry, and Mostafa Rahimi Azghadi. 2023. 'MFLD-net: a lightweight deep learning network for fish morphometry using landmark detection', *Aquatic Ecology*, 57: 913-31.

Saleh, Alzayat, Marcus Sheaves, Dean Jerry, and Mostafa Rahimi Azghadi. 2022. 'Transformer-based Self-Supervised Fish Segmentation in Underwater Videos', *arXiv preprint arXiv:2206.05390*.

Sinha, Ashish, and Jose Dolz. 2020. 'Multi-scale self-guided attention for medical image segmentation', *IEEE journal of biomedical and health informatics*, 25: 121-30.

Sun, Liang, Wei Shao, Qi Zhu, Meiling Wang, Gang Li, and Daoqiang Zhang. 2023. 'Multi-scale multi-hierarchy attention convolutional neural network for fetal brain extraction', *Pattern Recognition*, 133: 109029.

Sun, Yongheng, Duwei Dai, Qianni Zhang, Yaqi Wang, Songhua Xu, and Chunfeng Lian. 2023. 'MSCA-Net: Multi-scale contextual attention network for skin lesion segmentation', *Pattern Recognition*, 139: 109524.

Sung, Hee-Jee, Myeong-Kwan Park, and Jae Weon Choi. 2020. 'Automatic grader for flatfishes using machine vision', *International Journal of Control, Automation and Systems*, 18: 3073-82.

Tao, Andrew, Karan Sapra, and Bryan Catanzaro. 2020. 'Hierarchical multi-scale attention for semantic segmentation', *arXiv preprint arXiv:2005.10821*.

Thayananthan, Thevathayarajh, Xin Zhang, Wenbo Liu, Tianqi Yao, Yanbo Huang, Nuwan K Wijewardane, and Yuzhen Lu. 2023. "Automating catfish cutting process using deep learning-based semantic segmentation." In *Sensing for Agriculture and Food Quality and Safety XV*, 103-16. SPIE.

Usman, Muhammad, Azka Rehman, Sharjeel Masood, Tariq Mahmood Khan, and Junaid Qadir. 2024. 'Intelligent healthcare system for IoMT-integrated sonography: Leveraging multi-scale self-guided attention networks and dynamic self-distillation', *Internet of Things*: 101065.

Velásquez, Ricardo Arias, Jennifer Vanessa Mejía Lara, and Renato Arias Velásquez. 2023. "Evaluation of fruit selection with ensemble model in the Peruvian industry." In *2023 IEEE XXX International Conference on Electronics, Electrical Engineering and Computing (INTERCON)*, 1-6. IEEE.

Vo, Thi Thu Em, Hyeyoung Ko, Jun-Ho Huh, and Yonghoon Kim. 2021. 'Overview of smart aquaculture system: Focusing on applications of machine learning and computer vision', *Electronics*, 10: 2882.

Vu, Van Sang. 2022. 'Application of modern quantitative and molecular genetic tools in enhancing efficiency of Portuguese oyster (Crassostrea angulata) selective breeding in VIetnam'.

Wang, Libo, Ce Zhang, Rui Li, Chenxi Duan, Xiaoliang Meng, and Peter M Atkinson. 2021. 'Scale-aware neural network for semantic segmentation of multi-resolution remote sensing images', *Remote sensing*, 13: 5015.

Wang, Weizhen, Suyu Wang, Yue Li, and Yishu Jin. 2021. 'Adaptive multi-scale dual attention network for semantic segmentation', *Neurocomputing*, 460: 39-49.

Wang, Yongjun, Baiying Lei, Ahmed Elazab, Ee-Leng Tan, Wei Wang, Fanglin Huang, Xuehao Gong, and Tianfu Wang. 2020. 'Breast cancer image classification via multi-network features and dual-network orthogonal low-rank learning', *IEEE Access*, 8: 27779-92.

Wu, Xiaohong, Xinyue Liang, Yixuan Wang, Bin Wu, and Jun Sun. 2022. 'Non-destructive techniques for the analysis and evaluation of meat quality and safety: a review', *Foods*, 11: 3713.

Xie, Saining, Ross Girshick, Piotr Dollár, Zhuowen Tu, and Kaiming He. 2017. "Aggregated residual transformations for deep neural networks." In *Proceedings of the IEEE conference on computer vision and pattern recognition*, 1492-500.

Xu, Chunbo, Meng Lou, Yunliang Qi, Yiming Wang, Jiande Pi, and Yide Ma. 2021. 'Multi-scale attention-guided network for mammograms classification', *Biomedical Signal Processing and Control*, 68: 102730.

Xu, Xi, Yi Qin, Dejun Xi, Ruotong Ming, and Jie Xia. 2022. 'MulTNet: A Multi-Scale Transformer Network for Marine Image Segmentation toward Fishing', *Sensors*, 22: 7224.

Yu, Chuang, Yunpeng Liu, Zhuhua Hu, and Xin Xia. 2022. 'Precise segmentation and measurement of inclined fish's features based on U-net and fish morphological characteristics', *Applied Engineering in Agriculture*, 38: 37-48.

Zhang, Shanshan, and Hai Gao. 2021. "Application of Multi-network Fusion in Diagnosis of Chest Diseases." In *Proceedings of 2020 Chinese Intelligent Systems Conference: Volume I*, 358-66. Springer.

Zhao, Shili, Song Zhang, Jincun Liu, He Wang, Jia Zhu, Daoliang Li, and Ran Zhao. 2021. 'Application of machine learning in intelligent fish aquaculture: A review', *Aquaculture*, 540: 736724.

Zhou, Yuran, Qianqian Kong, Yan Zhu, and Zhen Su. 2023. 'MCFA-UNet: Multiscale cascaded feature attention U-Net for liver segmentation', *IRBM*: 100789.